# Tell me the truth: A system to measure the trustworthiness of Large Language Models

Dr. Carlo Lipizzi[1*]

[1] School of Systems and Enterprises, Stevens Institute of Technology, Hoboken, NJ, USA
* clipizzi@stevens.edu

## Abstract

Large Language Models (LLM) have taken the front seat in most of the news since November 2022, when ChatGPT was introduced. After more than one year, one of the major reasons companies are resistant to adopting them is the limited confidence they have in the trustworthiness of those systems. In a study by (Baymard, 2023), ChatGPT-4 showed an 80.1% false-positive error rate in identifying usability issues on websites. A Jan. '24 study by JAMA Pediatrics found that ChatGPT has an accuracy rate of 17% percent when diagnosing pediatric medical cases (Barile et al., 2024). But then, what is "trust"? Trust is a relative, subject condition that can change based on culture, domain, individuals. And then, given a domain, how can the trustworthiness of a system be measured? In this paper, I present a systematic approach to measure trustworthiness based on a predefined ground truth, represented as a knowledge graph of the domain. The approach is a process with humans in the loop to validate the representation of the domain and to fine-tune the system.

Measuring the trustworthiness would be essential for all the entities operating in critical environments, such as healthcare, defense, finance, but it would be very relevant for all the users of LLMs.

## Introduction

Trusting a system such as an LLM involves decision-making, being the decision of trusting or not trusting the system a choice the trustor needs to make. The trustor evaluates the risk of making a decision in this case by considering factors such as the system's capabilities, reliability, and the potential risks versus benefits of relying on its outputs.

As supported by vast literature (for example (Holsapple, 2008)), we base our decision on knowledge. This knowledge can be from direct experience or from sources we rely on. Sources can be humans - such as subject matter experts - or from tools we "trust".

A range of studies have explored the use of knowledge tools in decision-making. (Boose et al., 1992) and (Brown, 1989) both emphasize the importance of comprehensive decision models, with Boose specifically advocating for the use of knowledge acquisition techniques. (H. Kopackova et al., 2007) and (Rundall et al., 2007) focus on preprocessing tools and evidence-informed decision-making, respectively, with Rundall proposing the Informed Decisions Toolbox.

(Ytsen Deelstra et al., 2003) and (Yost et al., 2014) both stress the need for a strong connection between knowledge and decision-making, with Yost providing a practical example of tools used in public health departments. (Yavuz et al., 2005) takes a more practical approach, using a knowledge-based system to develop a marketing decision model. Collectively, these studies highlight the potential of knowledge tools in enhancing decision-making processes.

The method I present in this paper is a novel approach to measuring the trustworthiness of a Large Language Model and is a combination of Large Language Models, Knowledge Graphs and RDF triplets (subject, predicate, object), with humans in the loop.

The prototype I developed shows encouraging results.

## The overall context

There is an increasing interest and curiosity around "Artificial Intelligence". After ChatGPT was announced in November 2022, there has been a further acceleration. Most general news and articles have removed the lines of demarcation between AI, Natural Language Processing (NLP) and Machine Learning (ML). According to Google Trends, there have been over 10 million Google searches for ChatGPT since it was released in November 2022.

There were over 250,000 publications on AI in 2022, including journal articles, conference papers, other magazines and books (intelligence, 2022).

Companies are increasingly approaching Large Language Models. A study by Arize AI (*Survey: Large Language Model Adoption Reaches Tipping Point - Arize AI*, n.d.)shows an increase in adoption - or plans for adoption - in April '23 vs. September '23. One of the concerns that is growing the most is the lack of generic "accuracy" in the systems. If the answers are not "accurate", the system cannot be trusted.

## The literature

A range of studies have explored the measurement of trust in large language models (LLMs). (Koehl & Vangsness, 2023) highlight the potential of LLMs in analyzing qualitative responses. This is an interesting perspective, but it does not address the problem of getting a quantitative evaluation of bias in LLMs. (Yang Liu et al., 2023) focus on the importance of alignment with human intentions. (Yue Huang et al., 2023) have an interesting approach to evaluating LLMs in three crucial areas: toxicity, bias, and value alignment. Their approach ("TrustGPT") is focused on ethical and moral compliance of LLMs, which is a relevant one, but it leaves out the cases where "trust" overlaps with "accuracy" for a given domain or subject.

(L. Sun et al., 2024) is a massive study ("TrustLLM") conducted by 23 researchers defining trustworthiness for an LLM as a combination of truthfulness, safety, fairness, robustness, privacy, machine ethics, transparency, and accountability. They measure the answers provided by a given LLM against "gold answers" from given datasets. While this approach has evident merits, it does not address the problem of measuring the trustworthiness related to the specific domain the user is interested in.

Several other studies are primarily pointing to the problem without providing guidance for measuring the actual trustworthiness. For example, (P. Bhandari & H. M. Brennan, 2023) find that LLMs struggle to generate high-quality children's stories, and (Jacob Menick et al., 2022) emphasize the need for models to support answers with verified quotes. (Rick Rejeleene et al., 2024) have a more comprehensive approach, recognizing the importance of trust ("Trust plays a central role in economic transactions, for the majority of professions in businesses") and drilling down on the reasons why LLMs have trust issues. They base the evaluation of the quality of LLMs on three dimensions: accuracy, consistency and relevance. Their approach is theoretical, analyzing the logical/mathematical limitations of the algorithms and processes generating the LLMs. No actual measurement of the quality is provided.

Similarly to the previously cited authors, (Rachith Aiyappa et al., 2023) underscore the complexity of measuring trust in LLMs, analyzing the complexity of the process of creating an LLM and manually testing the accuracy of some LLMs to highlight the need for ongoing research in this area. They also cast a shadow on the existing methods to evaluate the performance of those models, often based on testing datasets that could be part of the datasets used to train the models, going against basic rules of data science (separation of training and testing) and common sense ("You can't judge a contest in which you are a participant").

Another element of complexity in the evaluation of trustworthiness is its subjectivity. While the fundamental concept of trust as a belief in the reliability and integrity of others (or systems) is a universal aspect of human life, the expression and significance of trust can vary widely across different cultures and contexts. (Rosanas, 2004) highlights the relevance of the "trustor" and the "trustee". Referring to the process of delivering trust, "The situation involves two decision makers, A (the trustor) and B (the trustee)". This echoes a previous position expressed by (Gambetta, 2000), who defines trust as "the subjective probability with which an agent expects that another agent or group of agents will perform a particular action on which its welfare depends".

## The goal of this paper

What seems to be lacking is a way to measure the trustworthiness of an LLM. Lord Kelvin in the 19th century (allegedly) said, "To measure is to know". That means we do not know if we can trust a system if we cannot have a measure of its trustworthiness.

Due to the difficulties of accounting factors such as common sense and logic, human evaluators are often involved in analyzing computer-generated output (Goodrich et al., n.d.). This is raising issues of consistency (Clark et al., n.d.) and scalability.

In this paper, I introduce a novel process to measure the trustworthiness of language - such as answers to questions - generated by an LLM. The process takes into consideration the subjectivity of the decision of trust and contemplates humans in the loop as subject matter experts of the domain. It is a quantitative approach aimed to measure how much a set of sentences provided as an answer to questions posted to an LLM is compatible with a given body of knowledge. This "compatibility" would act as a measurement of the trustworthiness of the answer provided by the LLM.

## The methodology

The system presented in this paper has two parts: the creation of the "truster" and its application.

### Building the truster

The following chart is a block representation of this subsystem.

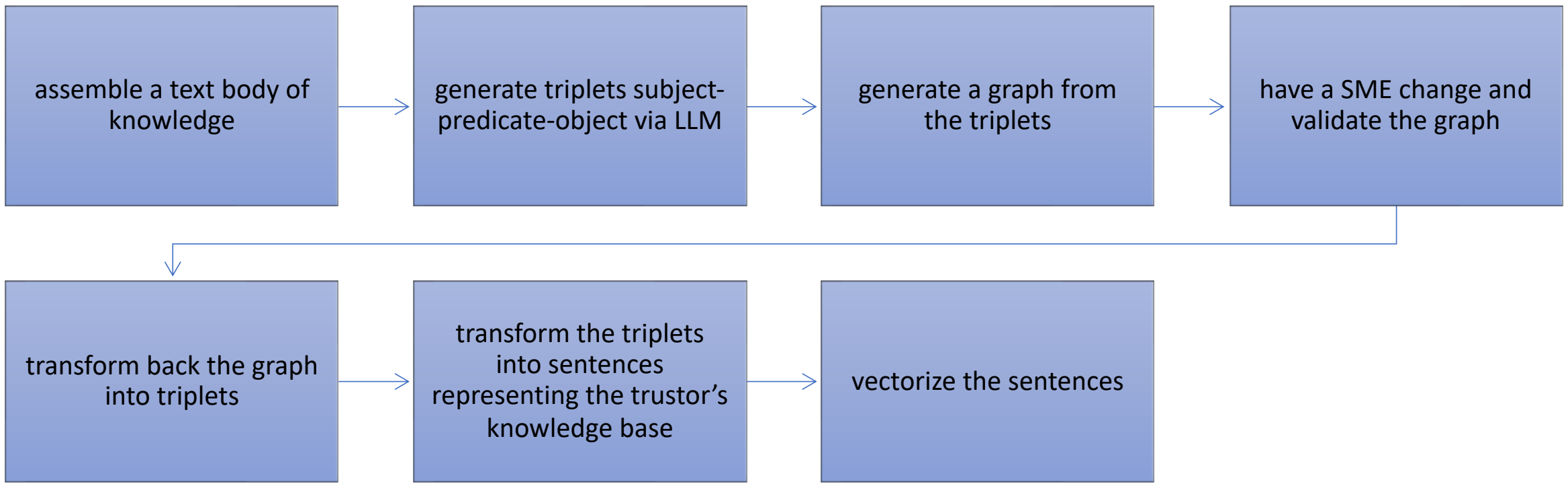

### Collecting the body of knowledge

As mentioned above, the process of "trust" is subjective and it is related to specific individuals, group, industry, culture. In this phase, a "representative" set of text is collected.

Being the whole system rooted in the same Machine Learning approach as LLMs, it is subject to the same limitations: underrepresented domains provide unsatisfactory results. The more focused the knowledge domain is, the easier it is to collect text sufficiently representing it.

On the opposite side of the spectrum, if the goal is a generic "trust", meaning a common sense-like trust, we should rely on datasets comparable with the ones used to train the leading LLMs. This would raise the issues of feasibility - due to the computational complexity - and of separation between testing and training, being the documents we would use likely part of the training set used for the LLM we are analyzing.

### Generating subject-predicate-object triplets

The goal of this phase is to represent the body of knowledge in a form that is simple yet powerful to model and understand information. The use of subject-predicate-object triplets is quite common in linguistics (such as in (Prasojo et al., 2018)). In this case, it also offers two benefits: making the text ready for encoding using the sentences generated by the triplets as tokens and using the triplets to create semantic graphs. Both the benefits will be detailed in the following paragraphs.

In order to generate the triplets, I used an LLM to create a first draft. Details in the case study.

### Generating a graph from the triplets

Up to this point, the process could have issues related to the representativeness of the text/body of knowledge and to the way the LLM transformed it into triplets. This is why a human subject matter expert should be involved to review and validate the results.

For as simple and powerful as the triplets' representation of the text could be, a long list of them may be difficult for a human to validate. Far easier to read and interact with is a directed graph where the subject and objects are nodes and predicates are edges.

### Having an SME editing and validating the graph

The triplets and the semantic graph extracted from them may have an inaccurate representation of the domain. This is why a human in the loop as an SME is introduced in this phase. Using GUI-based tools, humans can edit graphs easily. Once the SME is OK with the results, the edited graph can be saved and placed back in the process.

### Transforming back the graph into triplets

The final goal is to compare a representation of the domain/body of knowledge with the outcome of the LLM to evaluate its trustworthiness. The comparison could be made between text - via vectorization - or via graphs representing both sides. Comparing graphs has been done in different ways, some leveraging on the topological elements of the graph, such as in (Zhu & Iglesias, n.d.) and (Alkhamees et al., 2021), some using embeddings extracted from the graph, such as (Xiao et al., n.d.). I found both approaches do not provide enough semantic relevance to the similarity, but - as mentioned in the conclusions/next step - this can be revised.

Because of the above, I reverted the graph into triplets for further processing.

### Transforming the triplets into sentences

For the final goal of evaluating the LLM outcome versus the body of knowledge, I will use vector representations for both. By transforming the triplets into sentences, I can have the body of knowledge tokenized in a semantically relevant way. Generally speaking, the process of text tokenization for embedding generation is complex, with various challenges and potential solutions, all with limitations (Mohan et al., 2016). There are recent studies exploring new approaches, such as (W. Sun et al., n.d.), but it still remains a critical step at the foundation of machine learning approaches to language processing.

Sentences generated by triplets are tokens with semantic relevance and they could be a good foundation for embedding generation.

Once this step is completed, the body of knowledge is transformed into a set of sentences representing the given domain and vetted by a human SME. This is an evolution of a result I developed as "room theory" (Lipizzi et al., n.d.) before the LLMs' availability.

### Vectorize sentences

The sentences are then vectorized. For vectorization, I used a transformer-based model, which is currently considered to achieve good results in this task (Rahali & Akhloufi, 2023). The resulting embeddings are stored in a vector database. The model will be the same to vectorize the sentences to be analyzed, but it may be different from the one used to generate the triplets. The vectorized body of knowledge will act as a computational version of the knowledge base for the given domain.

## Building the validator

The following chart is a block representation of this subsystem.

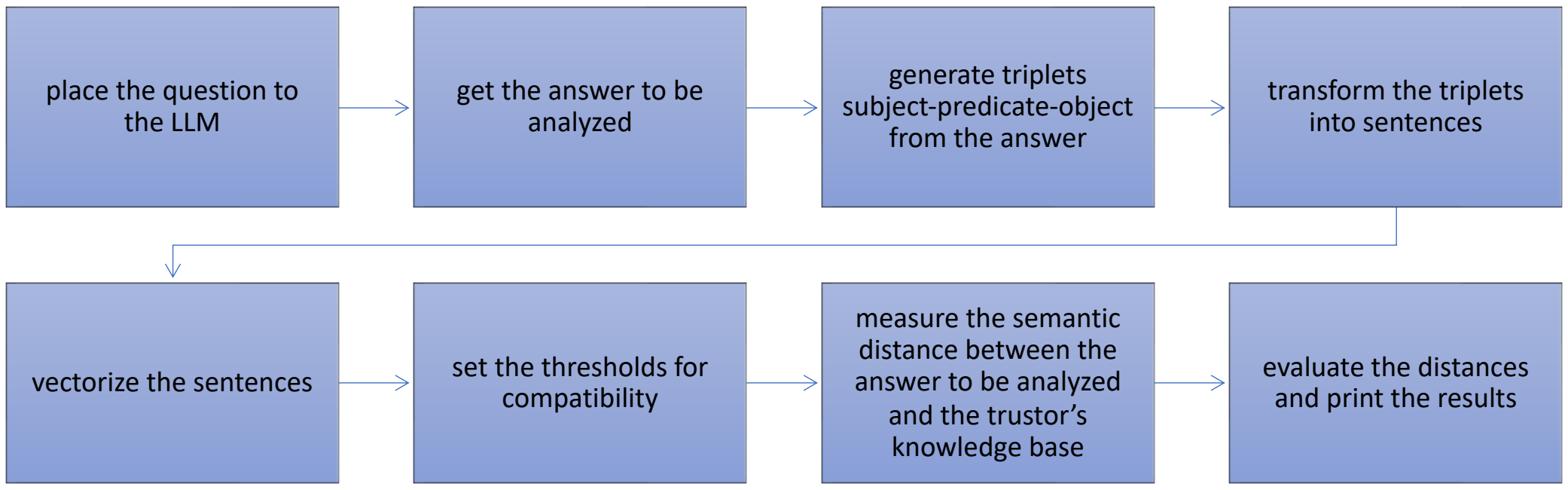

### Placing questions and getting answers from the LLM

The answer from the LLM is the one to be analyzed by the system to determine its trustworthiness, as depicted by the current body of knowledge.

### Generating subject-predicate-object triplets from the answer

The goal of this phase is to represent the answer in a way that is syntactically equivalent to the one used for the representation of the body of knowledge.

I used an LLM to generate the triplets. To avoid potentially different behaviors, I used the same model for the triplets' generation of the body of knowledge.

### Transforming the triplets into sentences and vectorize them

Like in the processing of the body of knowledge, triplets are transformed into sentences and sentences are then vectorized using the same model used to vectorize the body of knowledge. Due to the limited size of the embeddings in this case, they can be handled in memory without requiring vector databases.

### Set the thresholds for compatibility

This is a key step that has been addressed for this study only empirically, but it will be subject to further needed developments. The answer from the LLM may contain multiple sentences. Each one can or cannot be compatible with the body of knowledge. The compatibility of the individual sentences contributes then to the compatibility of the whole answer. That means that two thresholds need to be set: at the individual sentence level ($t_1$) and at the whole set of sentences making the answer ($t_2$).

### Measuring the semantic distances

The measures have the scope of estimating the distance between the vectors representing the answer and the vectors representing the body of knowledge. There are several ways to measure distance between vectors (Zhao, 2009). For this study, a plain cosine similarity was used as a measurement embedded in the specific vector database. This is another critical step, and it will be the subject of future developments. For example, in similar past studies I conducted, variations of Word Mover's Distance (Kusner et al., 2015) provided some advantages to cosine similarity.

### Evaluate the distances and print the results

Once the thresholds have been set, the system calculates the distance between each sentence and each of the sentences in the body of knowledge, printing those matching for values at or above the set threshold $t_1$. A summation of the individual scores creates the overall answer compatibility, which is then compared with the overall compatibility threshold $t_2$. The system will then print the result of the analysis in terms of compatibility or not. In case of non-overall compatibility, but in the presence of some sentence-level compatibility, a separate message will be printed.

## The proof of concept

A simple case study has been developed to test the applicability and validity of the methodology. Being a proof of concept, it has not the breadth and depth of a real application that will be developed in a future study.

### The body of knowledge

A relatively short description of "supply chain" has been used for this proof of concept. The text consists of about 400 words.

### The extraction of the triplets

To extract the triplets, I used GPT-4 via OpenAI API. This is another element that will be subject to revision, as well as testing different LLMs. I used the API in "chat-completion" mode. In this modality, I provided one text file with the prompt for the "assistant" and one for the "user". The "assistant" instructed the system to "extrapolate as many relationships as possible" in the form of [ENTITY 1, RELATIONSHIP, ENTITY 2], asking for a JSON format. The "user" was the text with the description of the body of knowledge related to the supply chain. The results have been preprocessed with basic NLP steps and placed in a CSV file for future reference. The triplets look like "supply chain, includes, sourcing".

### Generate and validate the graph

Using the triplets, I generated the edge list that is used to create the graph representing the given body of knowledge. The process is done using Python/NetworkX and saved in GML format. The GML file has then been imported into Gephi - an Open-Source graph editor - for editing. For this proof of concept, no SME has been involved, and no structural editing has been done on the graph. Thanks to its easy user interface, making changes to the graph could be done easily by an SME with no graph technical knowledge. The graph looks like the following:

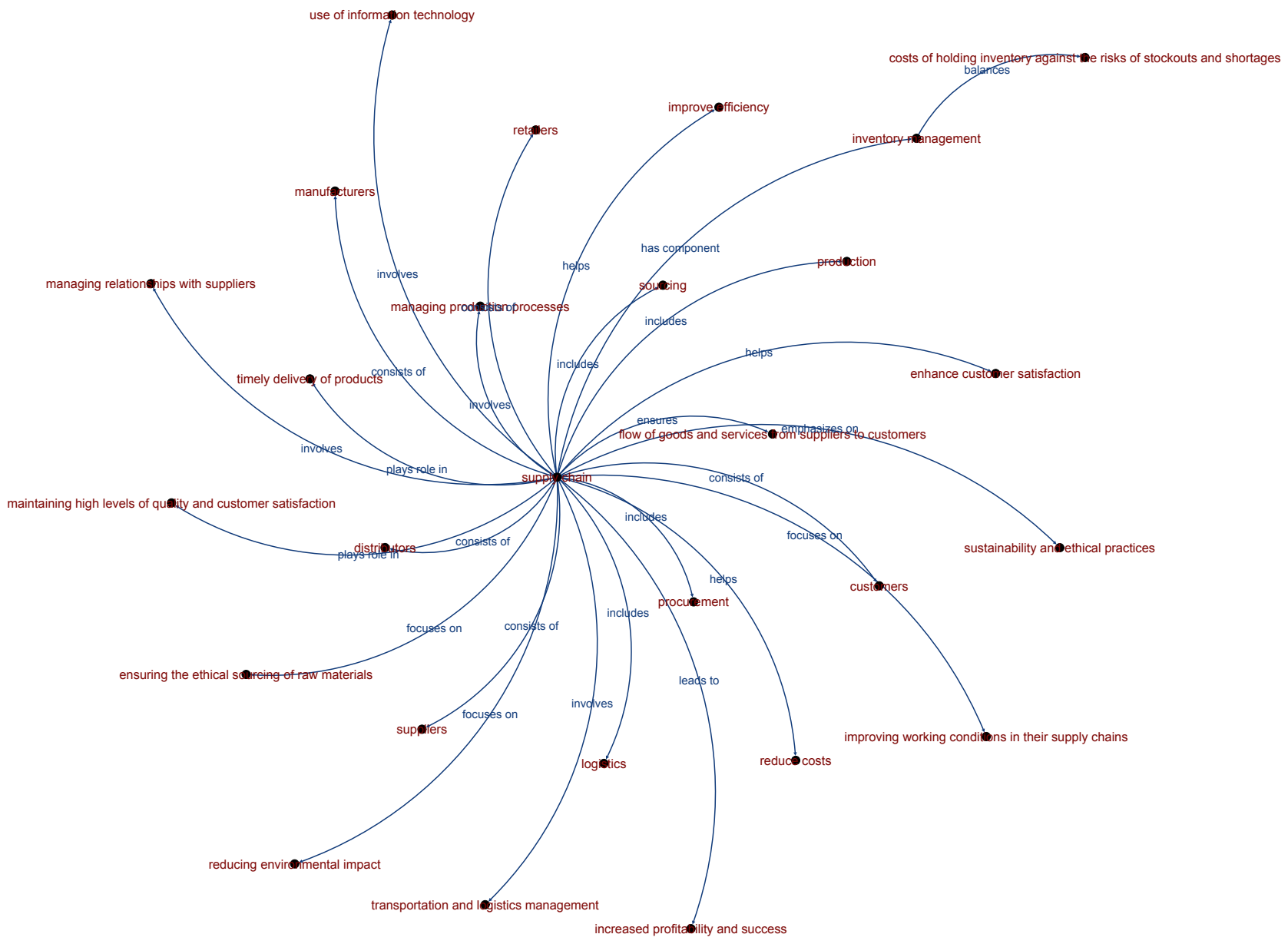

## Transforming the triplets into sentences and vectorize them

Because no change has been made to the graph, I used the same triplets that are creating the graph. If this would not have been the case, a further step graph->triplets would have been required. The transformation of triplets to sentences has been done in Python, generating sentences like "supply chain includes sourcing". For this proof of concept, I used the all-roberta-large-v1 model to vectorize the sentences and placed the resulting embeddings in a Pinecone vector database.

## Placing questions and getting answers

For this proof of concept, I used a set of simulated answers for the question/prompt "tell me facts about supply chain". The set of answers has been created as a combination of "reasonable" and "nonreasonable" answers to test the system. The answers I used are:

*"suppliers provide money", "suppliers provide joy", "supply chain gives you stamina", "suppliers provide materials", "supply chain transforms the world", "inventory management is a waste of time", "supply chain plays football", "supply chain goes to the moon"*

## Generating triplets from the answer, transforming the triplets into sentences and vectorize them

For this proof of concept, the answers are already in plain sentence form. In real cases, the answer from the LLM should have gone through the same process described above for the body of knowledge. The sentences have been vectorized using the same all-roberta-large-v1 model and placed in an in-memory storage.

### Set the thresholds for compatibility

I empirically set the threshold for individual sentence level ($t_1$). To calculate the threshold for the whole set of sentences making the answer ($t_2$), I considered 1. the percentage of sentences to be matched for the query to be considered compatible with the body of knowledge (*a*, set empirically) and 2. the number of phrases in the body of knowledge (*b*). The threshold $t_2$ is then a linear function of $t_1$, *a* and *b*. As mentioned in the methodology, this step will be subject to further developments.

### Measure and evaluate the distances/ print the results

I used the Pinecone embedded cosine similarity to measure the distance between the "answers" and the sentences in the body of knowledge. The following is an example of a compatible answer:

```
---Query: suppliers provide materials
   Matched sentence: supply chain includes sourcing
    - score: 0.65
   Matched sentence: supply chain includes procurement
    - score: 0.62
   Matched sentence: supply chain consists of suppliers
    - score: 0.6
   ---> The semantic proximity of this phrase to the knowledge base is: 1.88
        That means the phrase is compatible with the knowledge base
```

This is a sample of partial compatibility:

```
---Query: suppliers provide money
   Matched sentence: supply chain includes sourcing
    - score: 0.64
   Matched sentence: supply chain consists of suppliers
    - score: 0.63
   ---> The semantic proximity of this phrase to the knowledge base is: 1.27
        That means the phrase is not compatible with the knowledge base
        but there is some minimal compatibility
```

This is a sample of non-compatibility:

```
---Query: supply chain plays football
  -- No match: the phrase is not compatible with the knowledge base
```

## Conclusions and Future Developments

The proposed system delivers an effective method to measure the compatibility with the given body of knowledge of a set of sentences provided as an answer to a given question.

This compatibility is a black-box solution to the LLM trustworthiness question. As mentioned in the literature review, trust comprises several different components, such as accuracy, reliability, and alignment with human values and social norms. While accuracy and reliability can be considered characteristics of this system, the alignment with human, social and domain-specific values are supposed to be embedded in the body of knowledge in this system. This implies the critical role played by the construction of the body of knowledge in the approach described in this paper. Having an SME as a human in the loop is a way to address this issue, but further analysis should be done.

There are also technical aspects that should be further explored. The models used at the different stages should be compared to other available options. The vectorization used for comparing the body of knowledge and the answer from the LLM is based on sentence vectorization. Vectorizing the graphs representing the two components would be another option that could be explored better. The metric used here to measure the distance between sets of vectors should be compared to others. The definition of compatibility thresholds should be algorithmically defined, while the current proof of concept still relies on an empirical base.

This approach, based on measuring compatibility, could be used for different purposes. For example, in compliance analysis, where the system could match federal or regional guidelines with internal procedures. In general, it could be applied to all the cases where measuring the compatibility of a document with a given body of knowledge may be a valuable decision point.